\documentclass[a4paper,twoside]{article}

\usepackage[utf8]{inputenc}

\usepackage{amsmath}
\usepackage{graphicx}
\usepackage{caption}
\usepackage{subcaption}
\usepackage{siunitx}
\usepackage{gensymb}
\usepackage{tikz, pgfplots}
\usepackage{hyperref}

\usepackage{epsfig}
\usepackage{subcaption}
\usepackage{calc}
\usepackage{amssymb}
\usepackage{amstext}
\usepackage{amsmath}
\usepackage{amsthm}
\usepackage{multicol}
\usepackage{pslatex}
\usepackage{apalike}
\usepackage{svg}

\usepackage{SCITEPRESS}

\begin{document}

\title{Marine vessel tracking using a monocular camera}

\author{\authorname{Tobias Jacob\sup{1}\orcidAuthor{0000-0002-8530-258X}, Raffaele Galliera\sup{1}\orcidAuthor{0000-0001-7777-3835}, Muddasar Ali\sup{1}\orcidAuthor{0000-0002-6044-9892}, and Sikha Bagui\sup{1}\orcidAuthor{0000-0002-1886-4582}}
\affiliation{\sup{1}Department of Computer Science, University of West Florida, Pensacola, Florida, USA}
\email{\{tj75, rg101, ma229\}@students.uwf.edu, bagui@uwf.edu}
}

\abstract{In this paper, a new technique for camera calibration using only GPS data is presented. A new way of tracking objects that move on a plane in a video is achieved by using the location and size of the bounding box to estimate the distance, achieving an average prediction error of 5.55m per 100m distance from the camera. This solution can be run in real-time at the edge, achieving efficient inference in a low-powered IoT environment while also being able to track multiple different vessels.}

\keywords{Camera Calibration, Monocular Object Tracking, Object Detection, Error Prediction, Edge-ml, Distance Estimation, YoloV5, Kalman Filters.}

\onecolumn \maketitle \normalsize \setcounter{footnote}{0} \vfill

\section{\uppercase{Introduction}}

Ship avoidance systems are typically based on radio signals. In this paper, a new technique for camera calibration using only GPS data is presented. To get a better position estimate from an object detector, the property that ships move on a plane is utilized. Also, a new way of tracking objects that move on a plane in a video is achieved by using the location and size of the bounding box to estimate the distance.

This approach has many possible use cases in edge computing. This includes tracking airplanes in airports, ships in harbors, cars in parking lots, or humans in shopping malls. It also demonstrates how to efficiently track one specific object from other similar instances of that object. This can play an important role in traffic systems. For example, it is possible to keep track of free spaces in a parking lot. This information could be utilized in various IoT applications helping citizens to find a free parking space. Another use case is smarter traffic lights, which could use a camera to react to oncoming traffic quicker. For this, all that is needed for camera calibration and training or evaluating the object detector is 33 minutes of videos and ground truth GPS positions or something comparable. The methods utilized in this paper are fast enough to be deployed on inexpensive edge computing devices that can be mass-produced.

\section{\uppercase{Related work}}

A rich background of GPS camera calibration papers is available, e.g.~\cite{Liu2010}, but to date, none of these works use just GPS data to determine lens distortion of a single camera. This paper's novelty introduces an algorithm that relies on data from a single camera without any prior information about the camera parameters. Research has also been done for monocular position tracking with a calibrated camera.~\cite{7759904} used a car-mounted front-facing calibrated camera and Hough Transformations~\cite{10.1145/361237.361242} to extract the distance to other cars. They achieve an average lateral error of 9.71\%. The error is measured as the difference between prediction and ground truth in meters per 100 meters distance from the camera.~\cite{DBLP:journals/corr/abs-1809-10548} presents a combination between YoloV2 and a keypoint detector to achieve an average error of 6.25\%.~\cite{hu2019joint} uses Faster R-CNN for object detection and an additional network for 3D Box estimation, reaching an average position error of 7.4\%. Monocular position estimation is closely related to depth estimation. Recent methods achieved between 7.1\% and 21.5\% accuracy~\cite{Amlaan2019}. 

Some solutions have been proposed for Unmanned Aerial Vehicles (UAVs). \cite{9071439} suggests an approach to detect and estimate the location of an object in an environment where the UAV operates with a single monocular camera using its GPS position. Given the home position, the GPS location of the UAV, and the image taken from the camera, the authors have developed a method to estimate the depth of an object in the image after detecting it using Yolo. \cite{leira2017object} Developed a solution for UAVs with computational power and an onboard camera to detect, recognize, and track multiple objects in a maritime environment. The system was able to automatically detect and track the position and the velocity of a boat with an accuracy of 5-15 meters, being at the height of 400 meters. Furthermore, the tracking could be done even when a boat was outside the FOV (field of view) for extended periods of time.

In summary, several applications have been implemented to detect and track objects present in view of the camera, but not much work has been done on developing solutions that could run at the edge, leveraging low-powered IoT commercial off-the-shelf devices, which this work presents.

\section{\uppercase{Methods}}

The data consists of videos, a time series of GPS positions of the vessel, and the GPS position of the camera. First, the relation between the world position of the vessel and the on-screen position is determined. Then, an object detector is trained to track the vessel on the image. Post-processing is applied to increase stability and precision.

\subsection{Converting World / Screen}\label{sub:ConvertingWorldScreen}

Let $(\varphi, \lambda)$ be the latitude and longitude in arc measure, respectively, of the ship, and similarly $(\varphi_0, \lambda_0)$ be the position of the camera. The camera is at the origin at the world coordinate system, while the ship is at $(x_w, y_w)$. The distance in meters can be obtained with an equi-rectangular projection~\cite{Snyder1987}:
\begin{align}
    x_w &= r (\lambda - \lambda_0) \cos \varphi_0\,, \\
    y_w &= r (\varphi - \varphi_0)\,,
\end{align}
with $r = \SI{6371}{\kilo\meter}$ being the earth radius~\cite{Moritz2000}. The ship is always at sea level, so $z_w = 0$ can be omitted. The following homo-graphic equation has to be solved:

\begin{equation}\label{eq:homogenic}
    z_p \begin{pmatrix}
        x_p \\ y_p \\ 1
    \end{pmatrix} = H \begin{pmatrix}
        x_w \\ y_w \\ 1
    \end{pmatrix}\,.
\end{equation}

\begin{figure*}[p]
    \centering
    \begin{subfigure}[b]{0.9\textwidth}
        \centering
        \includegraphics[width=0.8\linewidth, trim=0 1.3cm 0 1.5cm, clip]{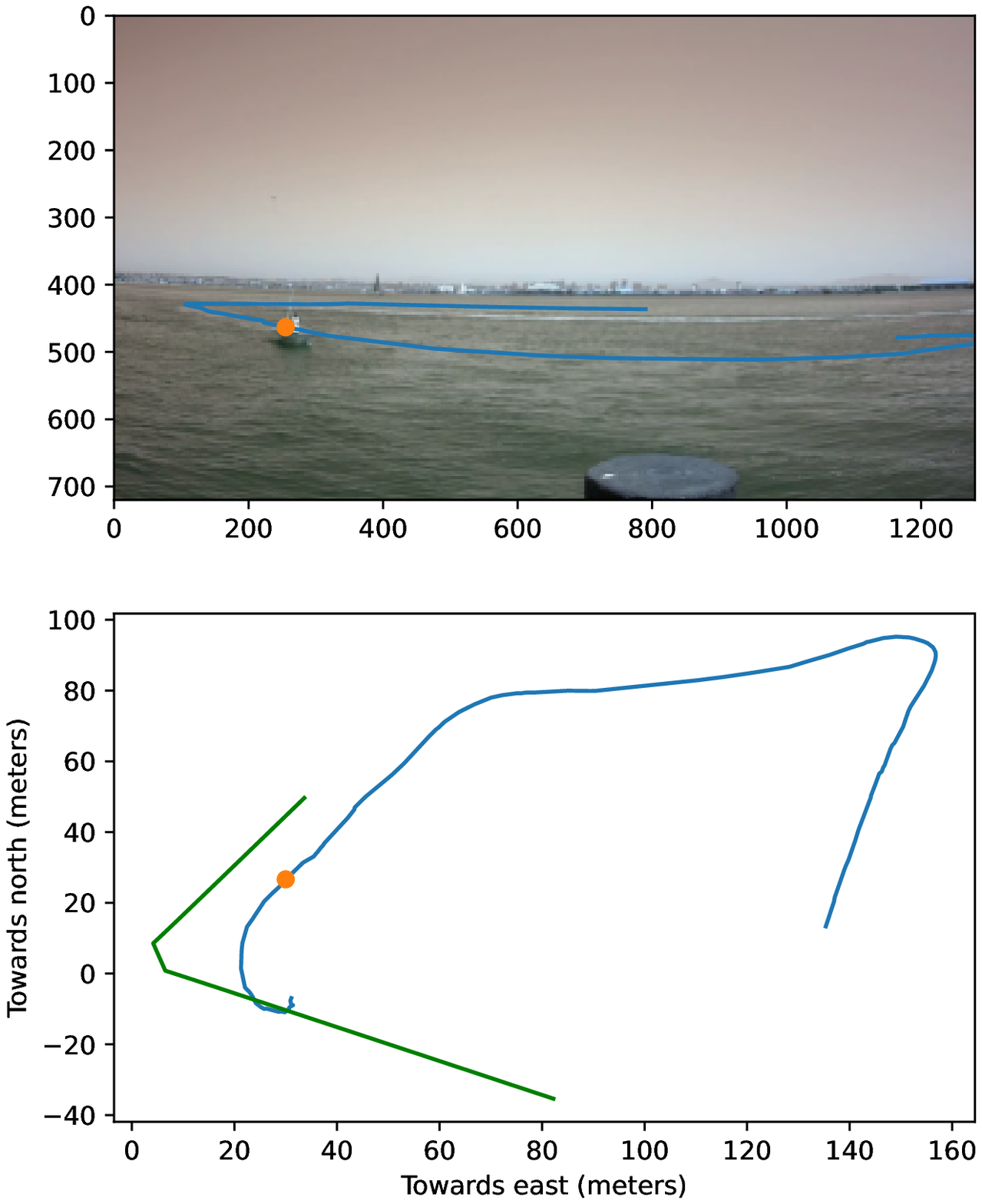}
        \caption{An example of a GPS track mapped into the video. Note that the camera is facing eastwards and not perfectly horizontal. The green lines indicate the rectangle $x_s \in [0, 1280], y_s \in [440, 720]$}
        \label{subfig:trajectoryProjected}
        \vspace{0.1cm}
    \end{subfigure}\vspace{1em}
    \begin{subfigure}[b]{0.6\textwidth}
        \centering
        \includegraphics[width=\linewidth]{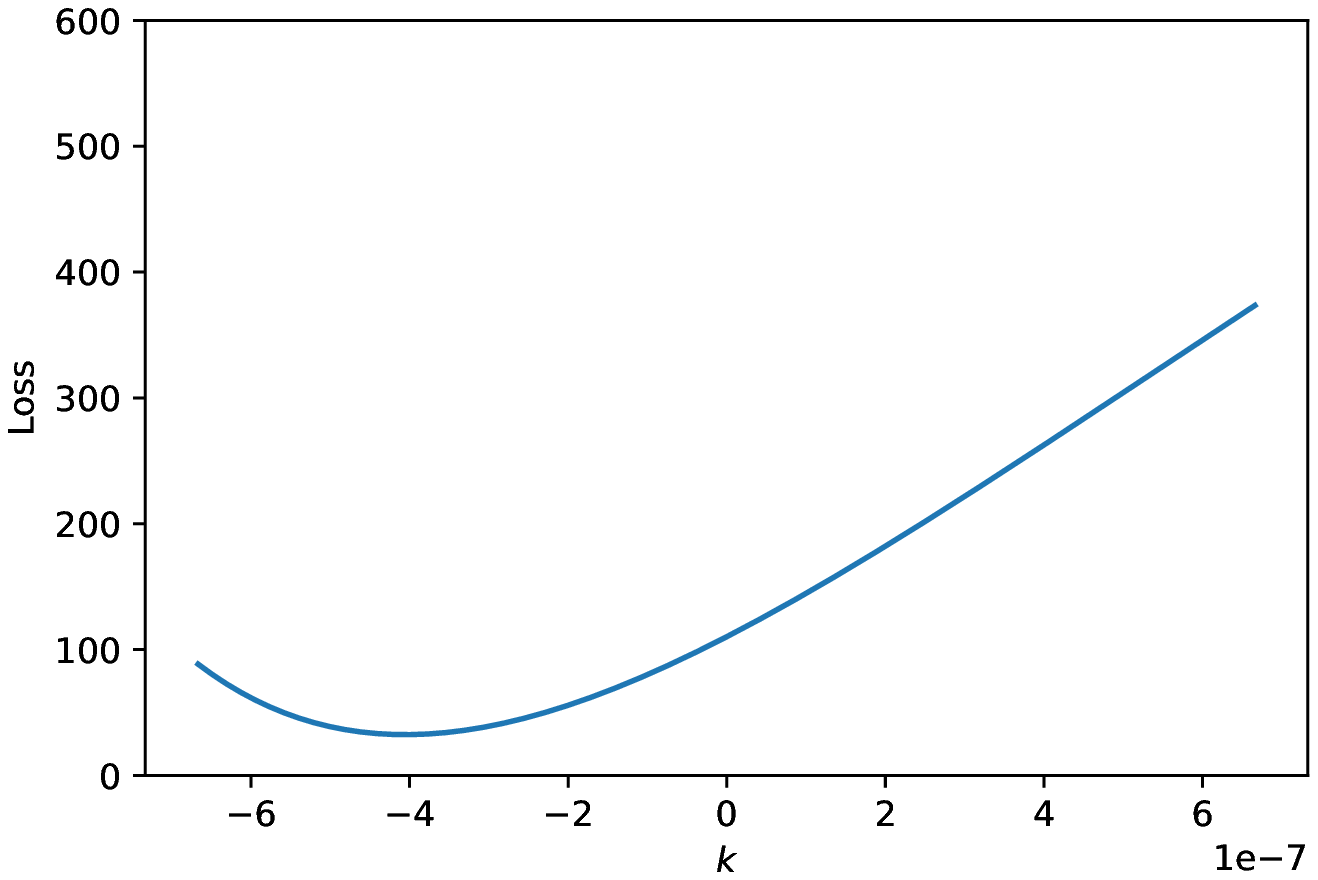}
        \caption{Brute force evaluation of the lens distortion parameter $K_1$}
        \label{subfig:lensDistortionEval}
        \vspace{0.1cm}
    \end{subfigure}
    \caption{Evaluation of the Camera Parameters}
    \label{fig:CameraParameters2}
\end{figure*}

$(x_p, y_p)$ are the projected coordinates. $z_p$ is the distance from the camera to the vessel. $H$ is a matrix that projects the world coordinates to the screen. The camera has significant lens distortion. This means that the final coordinate $(x, y)$ of the ship on screen and the projected coordinates $(x_p, y_p)$ have the relationship
\begin{align}
    x_p &= x_s + (x_s - x_c) (1 + K_1 r^2)\,, \\
    y_p &= y_s + (y_s - y_c) (1 + K_1 r^2)\,,
\end{align}
where $(x_c, y_c)$ is the center of the image and $r^2$ the distance from the center of the image.
\begin{equation}
    r^2 = (x_s - x_c)^2 + (y_s - y_c)^2\,.
\end{equation}
$K_1$ and $H$ are parameters that have to be learned. We created a dataset containing 11 hand-labeled pairs of screen positions $(x_{s, lab}, y_{s, lab})$ and GPS positions $(\lambda, \varphi)$. The GPS data was reconstructed from the video and frame index using a linear interpolation of the timestamps provided and a linear interpolation of the GPS data for that timeframe. The least-squares solution for equation~\ref{eq:homogenic} for $H$ can be easily computed. $K_1$ introduces a non-linearity; hence it was brute-forced. The result is shown in figure~\ref{subfig:lensDistortionEval}. Different values for $K_1$ were used for calculating $(x_{p, lab}, y_{p, lab})$ from $(x_{s, lab}, y_{s, lab})$ and solving equation~\ref{eq:homogenic} for $H$, and measuring the MSE between the estimated screen position $(x_p, y_p)$ and the hand-labeled screen position for all labeled points $i$.
\begin{equation}
    Loss = \sum_i (x_{p, i} - x_{p, lab, i})^2 + (y_{p, i} - y_{p, lab, i})^2\,.
\end{equation}

\subsection{Object detection}

\begin{table*}[t]
    \centering
    \begin{tabular}{c|c|c|c}
        Model & Image resolution & mAP (mean Average Precision) & Processed FPS \\
        \hline
        YoloV5x & 416x416px & 0.451 & 1.25\\
        YoloV5l & 416x416px & 0.328 & 2.48\\
        YoloV5m & 416x416px & 0.402 & 3.11\\
        YoloV5s & 416x416px & 0.333 & 8.70\\
        YoloV5s & 320x320px & 0.299 & 10.75\\
        \hline
        EfficientDet(D0) & 256x256px & 0.336 & 7\\
        EfficientDet(D1) & 256x256px & 0.394 & 5\\
    \end{tabular}
    \caption{Comparison of different YoloV5 and EfficientDet configurations, with mean Average Precision evaluated on a NVIDIA JetsonNano on COCO~\cite{lin2015microsoft} validation dataset and FPS performance reported on a 720p video. Models are sorted by type (Yolo, EfficientDet), size of the model, and adapted image resolution.}
    \label{tab:modelsComparison}
\end{table*}

The recent advancements of Deep Learning-based object detection models were state-of-art in achieving improvements in terms of both inference time and reliability of the predictions~\cite{8825470}. Dealing with the hardware constraints present in edge devices brings additional attention to the need for smaller models performing fast inference while keeping the accuracy of predictions as high as possible. We found two architectures and implementations,  YoloV5~\cite{glenn_jocher_2020_4154370} and EfficientDet~\cite{tan2020efficientdet} playing a key role in object detection model scenarios, offering different model sizes. We tested both models on a standard NVIDIA Jetson Nano Developer Kit, a small, commercial off-the-shelf device for AI embedded applications, equipped with 4GB of LPDDR4 RAM, 128-core Maxwell GPU, and a Quad-core ARM A57 CPU. 

As presented in table~\ref{tab:modelsComparison}, YoloV5s was the best solution for resource-constrained devices like the NVIDIA Jetson Nano regarding the trade-off between mAP (mean Average Precision) and FPS (Frames per Second) processed when using 320x320px images.

\begin{figure*}[p]
    \centering
    \includegraphics[width=.7\textwidth, trim=0 1.3cm 0 1.5cm, clip]{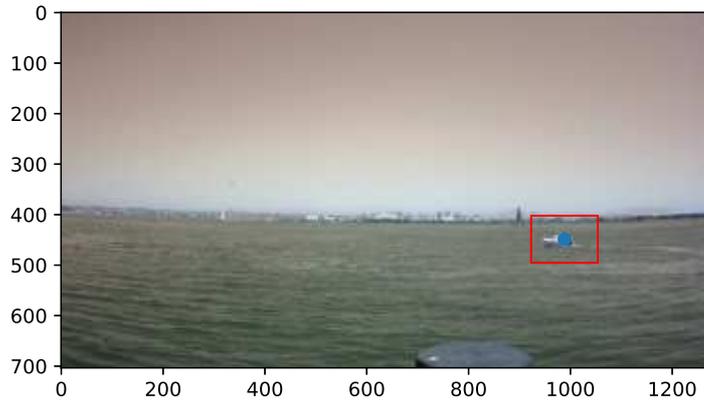}
    \caption{Example of bounding box size. It is empirically chosen to fit the boat.}
    \label{fig:exampleBoundingBox}
\end{figure*}

Once camera calibration and $H$ and $K_1$ are known, they can be used to predict the Bounding Box $(c_x, c_y, s_x, s_y)$ of the vessel. Let $(c_x, c_y)$ be the location of the ship. The dimensions of the bounding box $(s_x, s_y)$ are set on half image resolution to:
\begin{align}\label{eq:BBToZ_P}
    s_x = (1500 / z_p + 25)\,, \\
    s_y = (850 / z_p + 25)\,.
\end{align}
$z_p$ is the distance from the vessel to the camera in project space, as obtained by equation~\ref{eq:homogenic}. The scale factors were empirically chosen to include the vessel and have a minimum size of 25. An example is shown in figure~\ref{fig:exampleBoundingBox}. The image is scaled down by 2 and cropped to the upper $(640, 192)$ pixels to not process unnecessary data. Besides the hyper-parameters shown in table~\ref{tab:Hyperparameters}, the default hyper-parameters of YoloV5s were used.
\begin{table}[h]
    \centering
    \begin{tabular}{c|c}
        Epochs & 20 \\
        Batch size & 16 \\
        Image size & 352
    \end{tabular}
    \caption{Training hyper-parameters that differ from the default values.}
    \label{tab:Hyperparameters}
\end{table}

\subsection{Post-processing}

Outputs of the networks tend to be noisy. They might detect no ship, a slightly off position, multiple bounding boxes for the same ship, or multiple ships.

First, the trajectories for every ship have to be obtained. Each trajectory uses a Kalman filter~\cite{Kalman1960} to predict the location in the next frame. All detected ships in the next frame are accounted to the closest trajectory if they are within 80 pixels of the radius. Each trajectory is updated with the closest point that was accounted for it. A trajectory is finished if it is not updated for more than 10 frames. In case no trajectory is found for a point, it is considered new.

Out of all trajectories, the correct one has to be found. The trajectory score is defined as
\begin{equation}
    score = c_{0.8l} \log l\,.
\end{equation}
$c$ is the sorted confidence values that YoloV5s issued to the trajectory positions, with $l$ being the trajectory length for all data points $i$ of the trajectory.
\begin{equation}
    l = \sum_{i = 1}^{l - 1} \sqrt{(x_{w, i} - x_{w, i + 1})^2 + (y_{w, i} - y_{w, i + 1})^2}\,.
\end{equation}
$c_{0.8l}$ represents the 80th percentile of $c$. It was assumed that the trajectory with the highest score was the vessel we were looking for. When the trajectory got cut in the middle, other trajectories were pre- or appended to the current trajectory by looking at the timestamps.

\begin{figure*}[p]
    \centering
    \includegraphics[width=.6\textwidth]{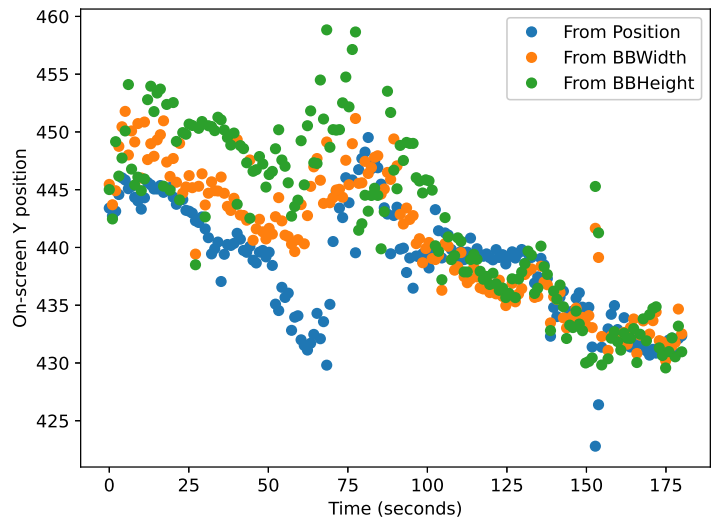}
    \caption{The $y_p$ by the two bounding-box-size-based estimators as in equation \ref{eq:BBToY_P} and the bounding-box-center-based predictor.}
    \label{fig:estimatingDistanceFromBoundingBoxAndPos}
\end{figure*}

\begin{figure*}[p]
    \centering
    \begin{subfigure}[b]{0.47\textwidth}
        \centering
        \includegraphics[width=\textwidth]{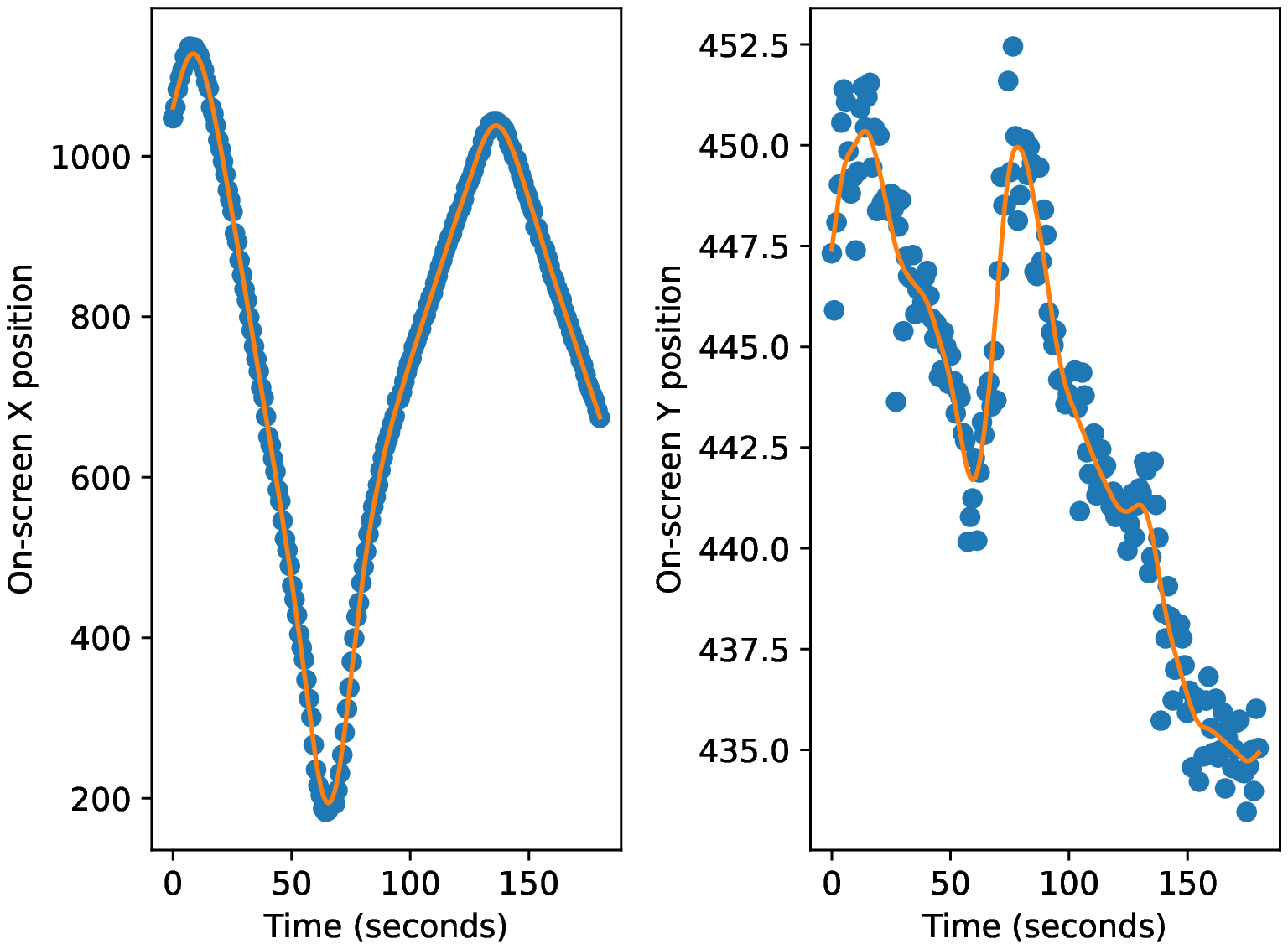}
        \caption{Effect of the Kalman filter for smoothing the trajectory in screen coordinates.}
        \label{subfig:KalmanSmoothingScreen}
        \vspace{0.3cm}
    \end{subfigure}\hspace{.05\textwidth}%
    \begin{subfigure}[b]{0.47\textwidth}
        \centering
        \includegraphics[width=\textwidth]{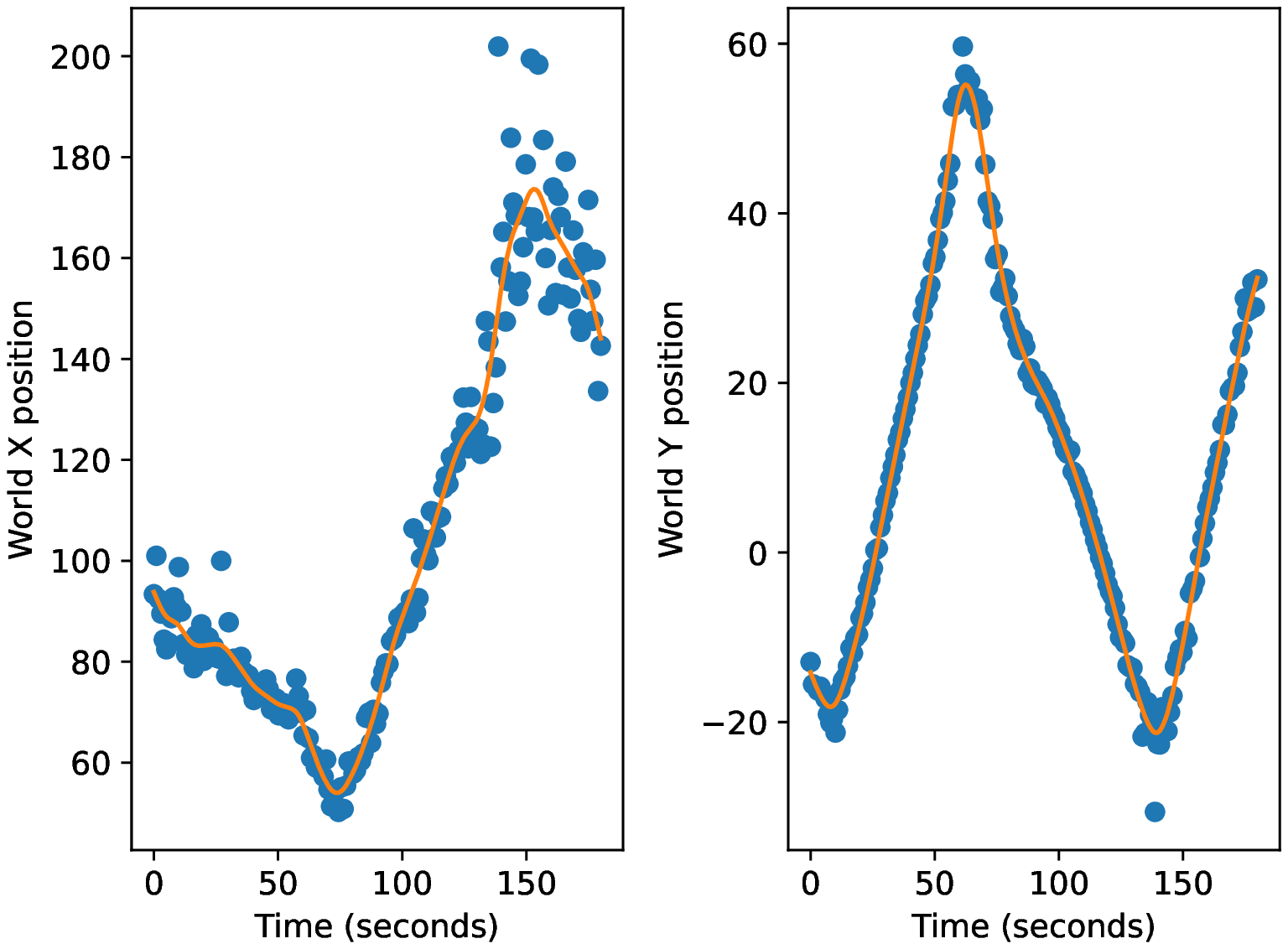}
        \caption{Effect of the Kalman filter for smoothing the trajectory in world coordinates.}
        \label{subfig:KalmanSmoothingWorld}
        \vspace{0.3cm}
    \end{subfigure}
    \caption{The Kalman filters are a crucial step in obtaining a smooth trajectory, particularly for lateral movement.}
    \label{fig:KalmanSmoothing}
\end{figure*}

$x_{p, Yolo}$ and $y_{p, Yolo}$ can be obtained directly from the object detector and camera calibration. The bounding box size is an additional indicator for distance $z_p$ and vertical position $y_p$. $z_{p, s_x}$ and $z_{p, s_y}$ can be obtained from equation~\ref{eq:BBToZ_P}.
\begin{equation}~\label{eq:BBToY_P}
    y_{p} = \frac{\pm \frac{1}{z_p} - H_{2, 0}^{-1} x_p - H_{2, 0}^{-1}}{H_{2, 1}^{-1}}\,.
\end{equation}
The different predictors are shown in figure~\ref{fig:estimatingDistanceFromBoundingBoxAndPos}. The final estimate of the y position is the weighted average:
\begin{equation}
    y_{p} = 0.5 y_{Yolo} + 0.25 y_{p, s_x} + 0.25 y_{p, s_y}\,.
\end{equation}
The parameters were empirically chosen since the position-based predictor is less noisy than the bounding box size-based predictors, as shown in figure~\ref{fig:estimatingDistanceFromBoundingBoxAndPos}. Finally, a Kalman filter is used to smooth the composed trajectory and estimate missing values. Points with a vertical difference of more than 5 pixels to the smoothed trajectory are considered outliers and masked out. Then the algorithm runs again, 5 times in total. A masked out point may become unmasked if the smoothed trajectory moved closer to it. The effect is shown in figure~\ref{fig:KalmanSmoothing}. Then, the GPS coordinate of the vessel can be obtained by reversing the equations in section~\ref{sub:ConvertingWorldScreen}. If the point is above the horizon ($z_p$ has the wrong sign), or very far, it is excluded. Missing data points may be obtained using a cubic interpolation of the data. The output is clipped to $[-2000, 2000]$ for safety.

All Kalman filters use the following state-transition $F_k$, observation $R_k$, transition co-variance $Q_k$ and observation co-variance $R_k$, with $\Delta$ being the time in seconds between frames:
\begin{align}
    F_k &= \begin{pmatrix}
        1 & 1 & 0 & 0 \\
        0 & 1 & 0 & 0 \\
        0 & 0 & 1 & 1 \\
        0 & 0 & 0 & 1
    \end{pmatrix}\,, \\
    H_k &= \begin{pmatrix}
        1 & 0 & 0 & 0 \\
        0 & 0 & 1 & 0 
    \end{pmatrix}\,, \\
    Q_k &= diag(0, 1 \Delta^2, 0, 1 \Delta^2)\,, \\
    R_k &= diag(10^2, 10^2)\,.
\end{align}

%------------------------------------------------

\section{\uppercase{Results}}

All methods were validated on the dataset of the ``AI tracks at sea'' Challenge~\cite{Tall2020}. This dataset contained 11 videos, each 3 minutes in time, and ground truth GPS positions for the ship.

\subsection{Camera Parameters}

\begin{figure*}[p]
    \centering
    \begin{subfigure}[b]{0.9\textwidth}
        \centering
        \includegraphics[width=\linewidth]{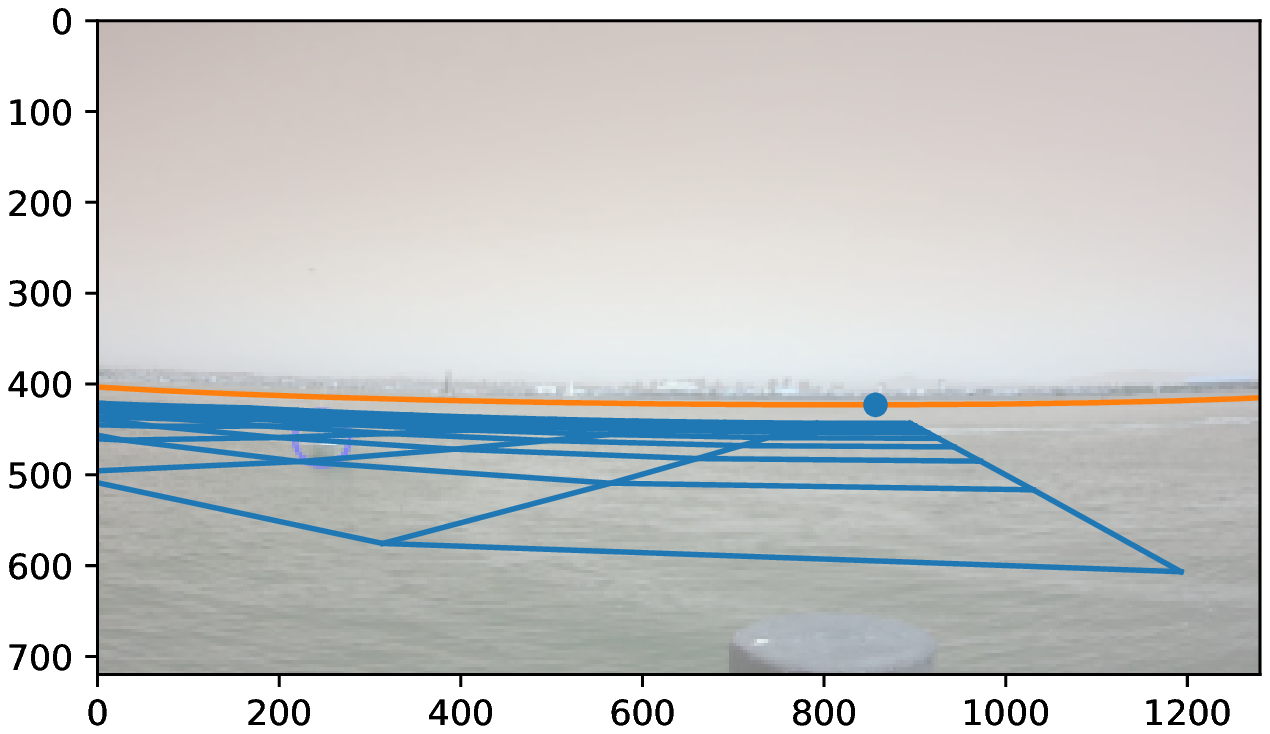}
        \caption{Visualization of the learned camera parameters. The grid shows 10 meter squares, with $100 > x_w > 10$ and $100 > y_w > 0$. Note the effect of the lens distortion, and how the orange line matches the horizon. The blue vanishing point is exactly east.}
        \label{subfig:projection}
        \vspace{0.1cm}
    \end{subfigure}
    \begin{subfigure}[b]{0.9\textwidth}
        \centering
        \includegraphics[width=\linewidth]{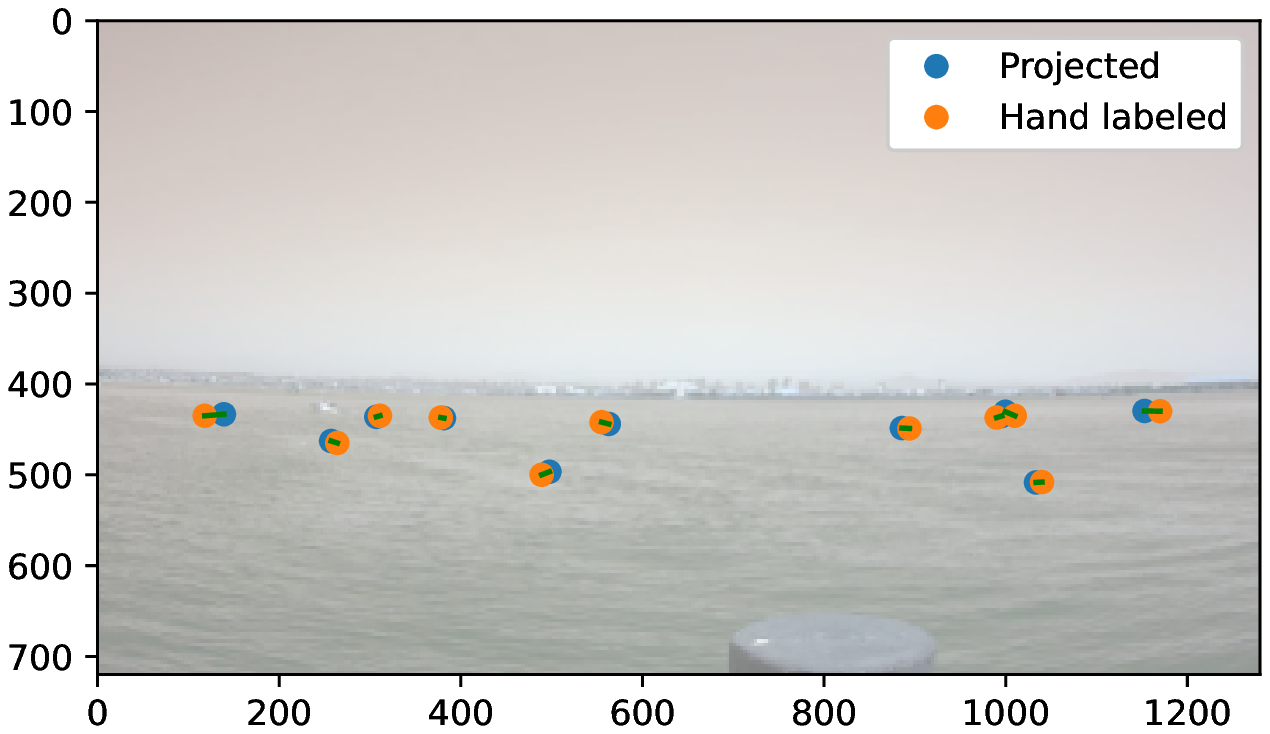}
        \caption{The difference between our hand labeled points and the projected GPS coordinates.}
        \label{subfig:projectionError}
        \vspace{0.1cm}
    \end{subfigure}
    \caption{Evaluation of the Camera Parameters}
    \label{fig:CameraParameters}
\end{figure*}

The learner for inferring the camera parameters
discovered the following K1 and H values:
\begin{footnotesize}
    \begin{align}
        K_1 &= -4.053 \cdot 10^{-7}\,, \\
        H &\approx \begin{pmatrix}
            4.41 \cdot 10^2 & 3.79 \cdot 10^2 & -1.00 \cdot 10^3 \\
            -2.17 \cdot 10^2 & -5.65 \cdot 10^1 & -5.48 \cdot 10^2 \\
            -5.12 \cdot 10^{-1} & -1.44 \cdot 10^{-1} & 1 \cdot 10^0
        \end{pmatrix}\,.
    \end{align}
\end{footnotesize}
The results are visualized in figure~\ref{fig:CameraParameters}. The orange line in figure~\ref{subfig:projection} is characterized by the vanishing points of the plane
\begin{equation}\label{eq:horizon}
    x_{p, hor} H^{-1}_{3, 1} + y_{p, hor} H^{-1}_{3, 2} = - H^{-1}_{3, 3}\,.
\end{equation}
It matches the actual horizon within 5 pixels. The difference between our hand-labeled points and the project points is shown in figure~\ref{sub@subfig:projectionError}. The loss is visible as the mean squared error between the orange and blue points, indicated by the green lines. The camera orientation can also be obtained. With $x_{p, hor} = x_c$ being in the middle of the image and $y_{p, hor}$ fulfilling equation~\ref{eq:horizon}, the horizon point can be deprojected using
\begin{equation}
    \begin{pmatrix}
        x_{w, hor} \\ 
        y_{w, hor} \\
        0 
    \end{pmatrix} = H^{-1} \begin{pmatrix}
        x_{p, hor} \\
        y_{p, hor} \\
        1
    \end{pmatrix}\,.
\end{equation}
The heading of the camera is
\begin{equation}
    \alpha = \arctan{\frac{x_{w, hor}}{y_{w, hor}}} = 76.5\degree\,.
\end{equation}
This allows for the adjustment of the camera's orientation by multiplying $(x_w, y_w)$ with a 2D rotation matrix.

\subsection{Object detection}

14 videos that contained recorded camera imagery of sea vessel traffic and the recorded GPS track of the vessel of interest were used for this work. The network was trained on 6 Videos (7, 8, 9, 10, 11, and 12). 5 videos (13, 14, 15, 16, and 17) were used for validation. Videos 18, 19, and 20 did not have GPS data, so they could only be used for testing. For final results, the network was trained for one additional epoch on the training and validation dataset. The results shown in this paper are without training on the validation set.

Two situations occurred depending on the training length. If YoloV5s was trained for 50 epochs, it distinguished the vessel from other ships like sailboats and jet skis on the training data, but it performed poorly on validation due to overfitting, which started after epoch 20. After 20 training epochs, YoloV5s not only detected the ship well but also other ships like sailboats and jet skis. Filtering the correct ship out of all the generated trajectories turned out to be a major source of error.

\subsection{Post-processing}

\begin{figure*}[p]
    \centering
    \begin{subfigure}{0.5\textwidth}
        \centering
        \includegraphics[width=\linewidth]{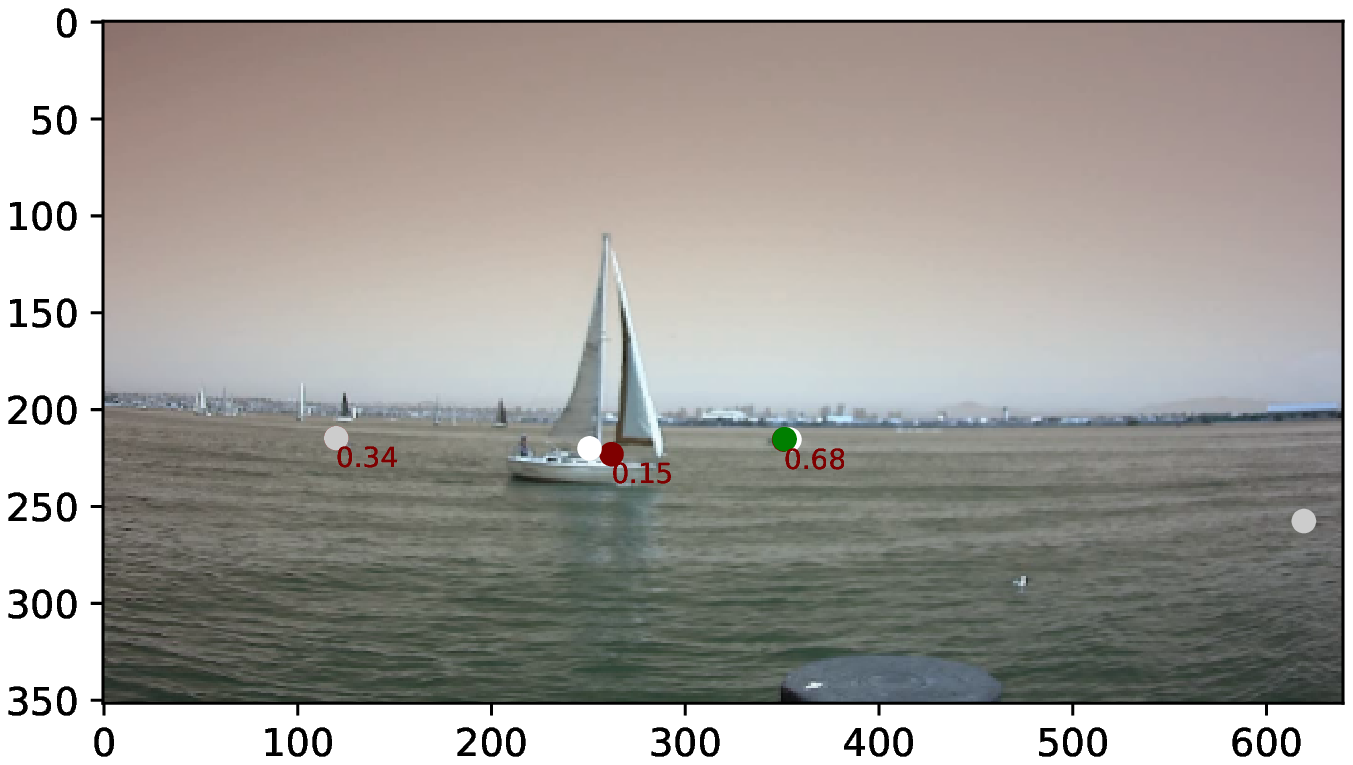}
    \end{subfigure}%
    \begin{subfigure}{0.5\textwidth}
        \centering
        \includegraphics[width=\linewidth]{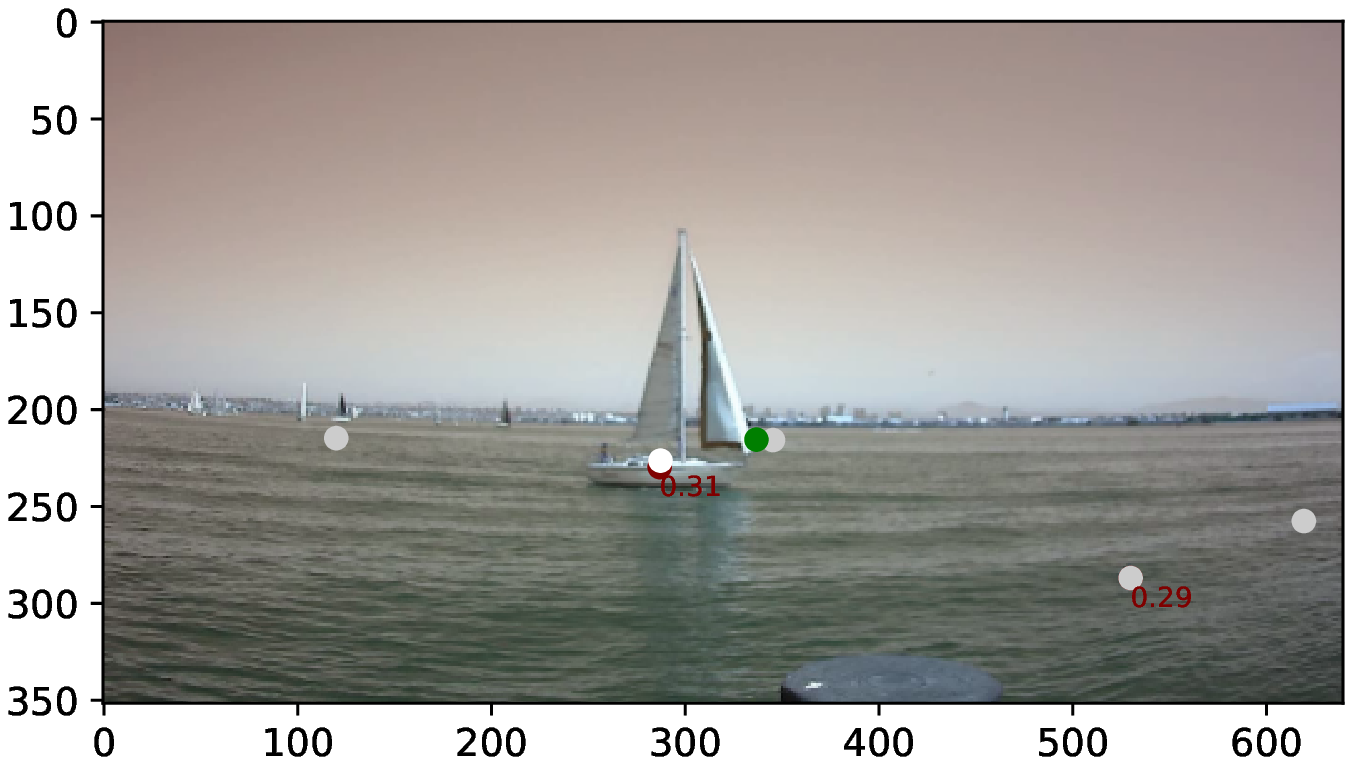}
    \end{subfigure}
    \begin{subfigure}{0.5\textwidth}
        \centering
        \includegraphics[width=\linewidth]{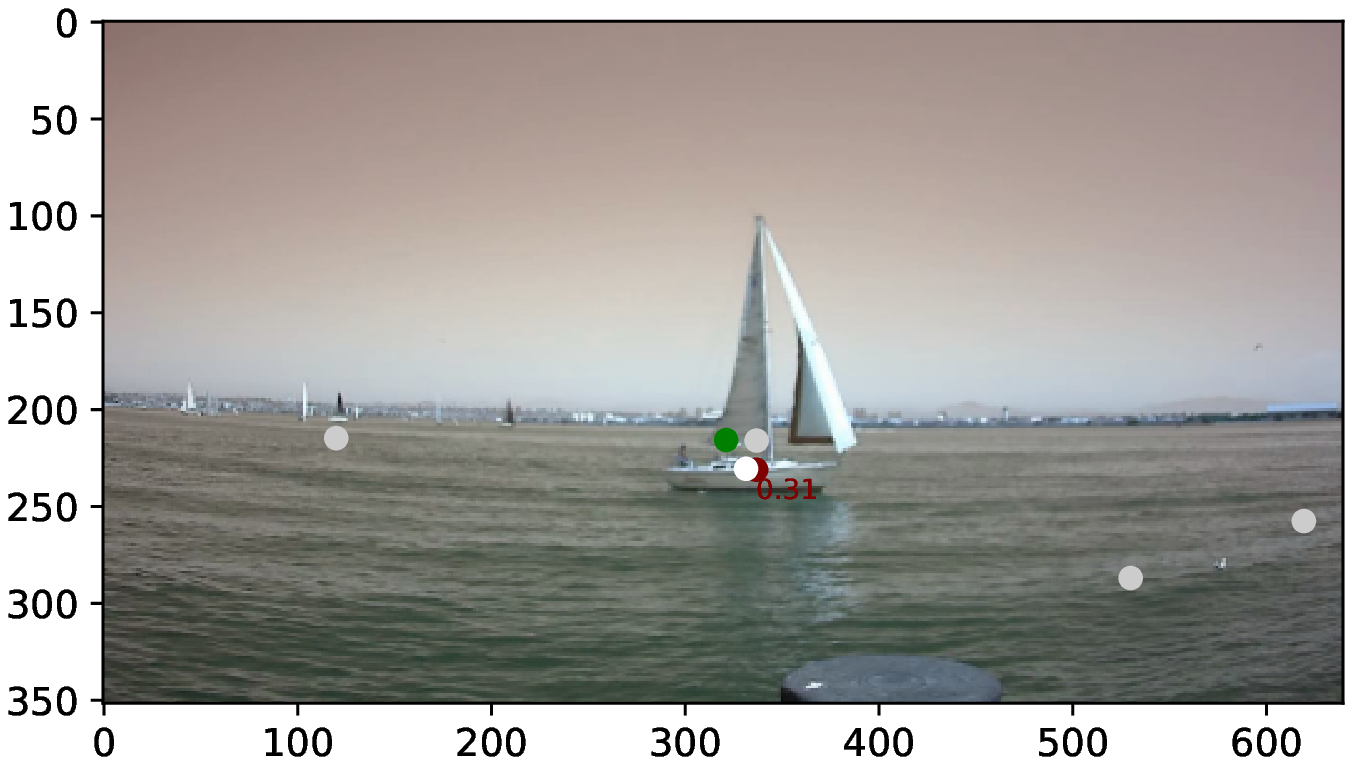}
    \end{subfigure}%
    \begin{subfigure}{0.5\textwidth}
        \centering
        \includegraphics[width=\linewidth]{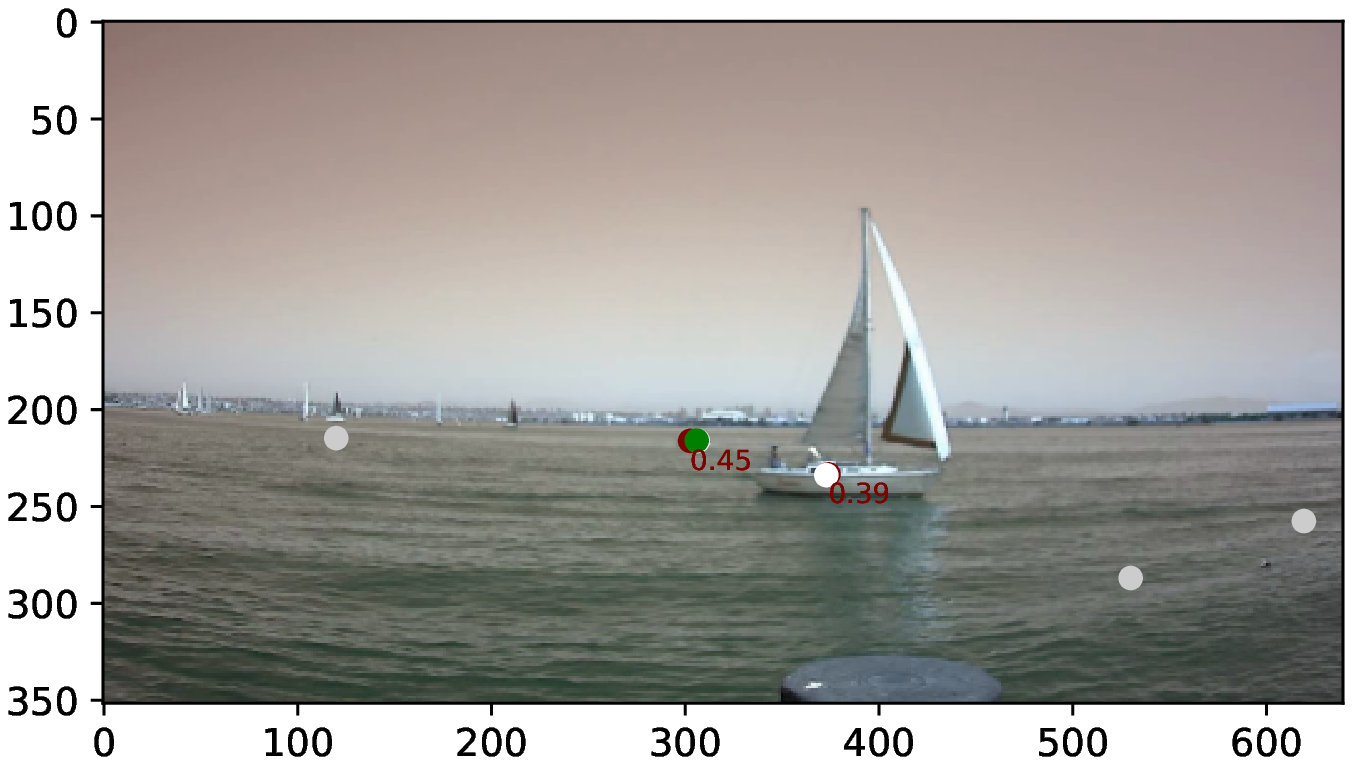}
    \end{subfigure}
    \caption{Green is the smoothed trajectory of the vessel. White/Grey is the trajectory of a ship in general (white when it got detected in this frame, grey if it has not been detected in the current frame and position is guessed). Red are detected points in the image. They may be obscured by the white points. When the vessel is behind the sailboat, the grey and green point estimate the position.}
    \label{fig:kalmanObscured}
\end{figure*}

YoloV5s infers a position that is 2 pixels (on half resolution) above the actual on-screen position needed. Therefore, this was added as a fixed offset to the positions that the network inferred. This could be attributed to the fact that the pre-trained YoloV5s used a different center point on objects. The Kalman filters can remember the velocity of each vessel. When two vessels cross each other, they can keep track of both trajectories, as shown in figure~\ref{fig:kalmanObscured}. In video 18, the model can interpolate the out-of-screen position of the vessel with the information of how the vessel left and reentered the frame.

\begin{figure*}[p]
    \centering
    \includegraphics[width=.80\linewidth]{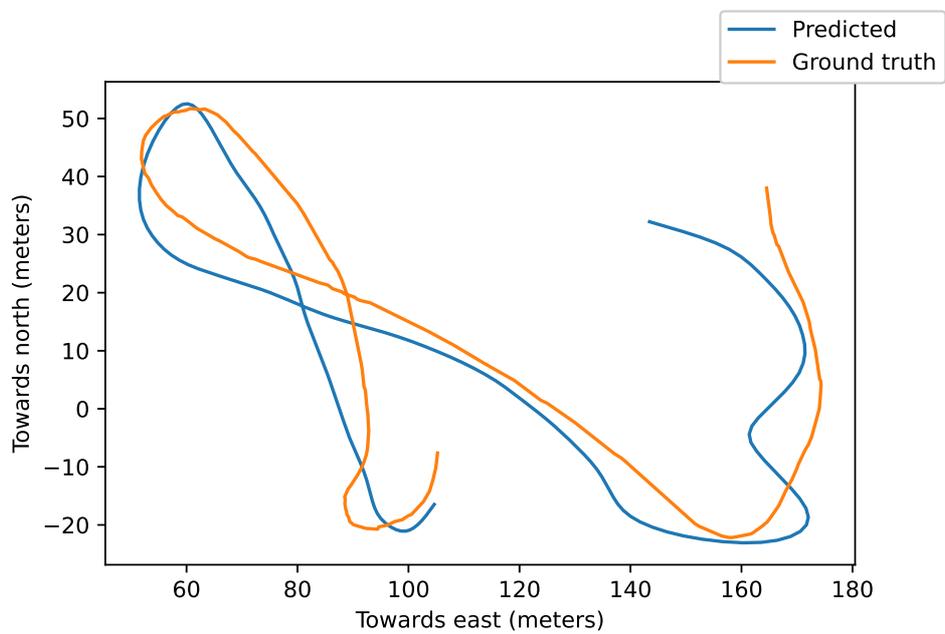}
    \caption{Validation of the complete pipeline using Video 15. Showing the Ground Truth track of the vessel (orange line) against the predicted one (blue line).}
    \label{fig:back2GPS}
\end{figure*}

The result of the final pipeline is shown in figure~\ref{fig:back2GPS} and figure~\ref{fig:errorDistribution}. The error for 4 out of 5 videos is less than $\SI{20}{\meter}$. The error is also proportional to the distance. At a distance of $\SI{150}{\meter}$, if the ship at $(x_w, y_w)$ moves roughly $\SI{10}{\meter}$ away from the camera ($(150, 50)$ to $(160, 53)$), this equates to a vertical change of roughly 0.276 pixels on the half resolution the network is working on. This approach is working with sub-pixel accuracy.

\subsection{Performance}

On the NVIDIA Jetson Nano used for testing, 5.31 FPS was reached for the whole pipeline. YoloV5s took 72\% of that time, while the other 28\% can be mostly attributed to video decoding. Considering that 1 FPS is enough for the algorithm to produce a stable output, this algorithm can easily run in real-time on low-powered embedded devices. 

\subsection{Validation}

The main objective is to validate performance by calculating the inference error of the model, which represents the distance in meters from the ground truth position against the distance in meters from the video source. A correlation between these two factors is valuable for understanding how the model performs. Additionally, the output will be enriched with the expected error, a great improvement of the pipeline, both from a technical and user-end perspective.

This section explains how to perform the distance estimation of the vessel from the camera. The distance of the predicted point from the ground truth will also be presented, ending with a brief explanation of possible error prediction methods.
As mentioned in the Object Detection section, the validation has been performed on five different videos (videos 13, 14, 15, 16, and 17) for a total amount of 713 data points.

\subsubsection{Distance estimation}
In order to calculate the distance from the camera and the actual error from the predicted point, the Haversine formula~\cite{MMGM2018} is used. The formula calculates the shortest distance between two points on a sphere using their latitude and longitude, and is expressed as follows:
\begin{equation}
    d = 2r\arcsin{\sqrt{\sin^2{\frac{\phi_{2} - \phi_{1}}{2}} + \cos{\phi_{1}}\cos{\phi_{2}}\sin^2{\frac{\lambda_{2} - \lambda_{1}}{2}}}}
\end{equation}
$r$ is the radius of the earth in meters, $\phi_{1}, \phi_{2}$ is the latitude of the two points, and $\lambda{1},  \lambda{2}$ is the longitude.

\subsubsection{Considerations about the validation set}
The distance from the camera of the target vessel is approximately between 86 and 200 meters for 75\% of the data points, with peaks of 2500 meters. However, one of the videos represents a particular issue for the validation process, as its frames show the vessel from the back and at a close distance, a point of view that is unique in this dataset. A possible solution could consist of shrinking the testing set by including the video in the training process. Although this work aims to get the best performance possible in terms of inference and generalization with a limited amount of data points, we found that the model cannot generalize to that level in this constrained scenario. 

\subsubsection{Validation results}

\begin{table*}[t]
    \centering
    \small
    \begin{tabular}{c|c|c|c|c}
        Paper & Method & Target Object & Image resolution & Avg. error \\
        \hline
        \cite{7759904} & Hough Transform & Cars & 480x360 & 9.71\% \\
        \cite{hu2019joint} & Faster R-CNN and 3D box prediction & Cars & 1392x512 & 7.40\% \\
        \cite{DBLP:journals/corr/abs-1809-10548} & YoloV2 and keypoint prediction & Cone & Not stated & 6.25\% \\
        Ours & YoloV5s & Vessel & 640x360 & 5.55\%
    \end{tabular}
    \caption{Comparison of tracking accuracy with others. The average error is measured in meters per 100 meters distance. Different cameras, angle of views, target objects, and image resolution make a direct comparison hard to value.}
    \label{tab:accuracyComparison}
\end{table*}

\begin{figure*}[p]
    \centering
    \includegraphics[width=0.9\textwidth]{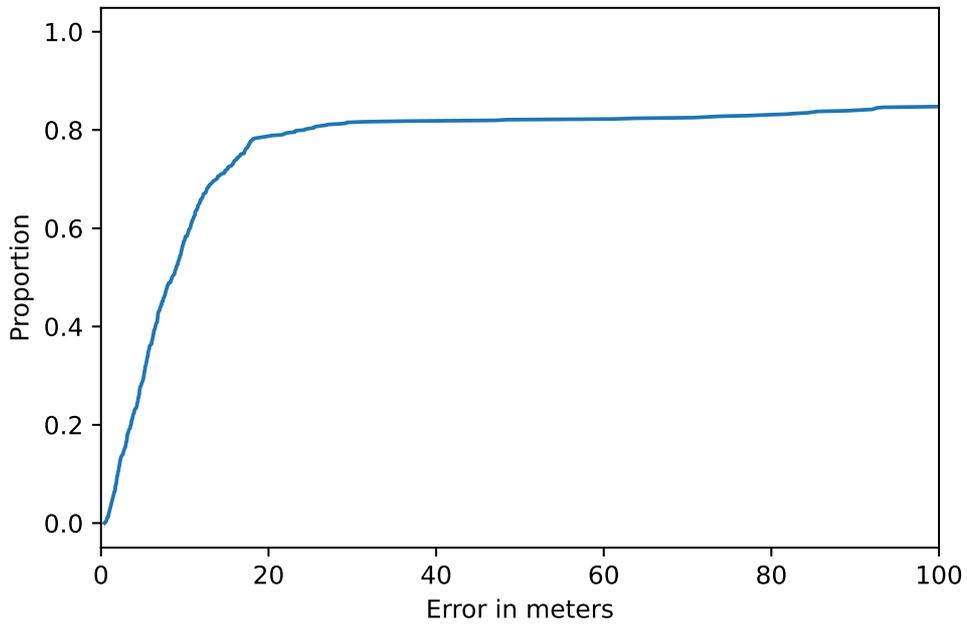}
    \caption{Distribution of the error in meters, output of our pipeline. 80\% of the predictions have an error below 20 meters.}
    \label{fig:errorDistribution}
\end{figure*}

\begin{figure*}[p]
    \centering
    \includegraphics[width=0.9\textwidth]{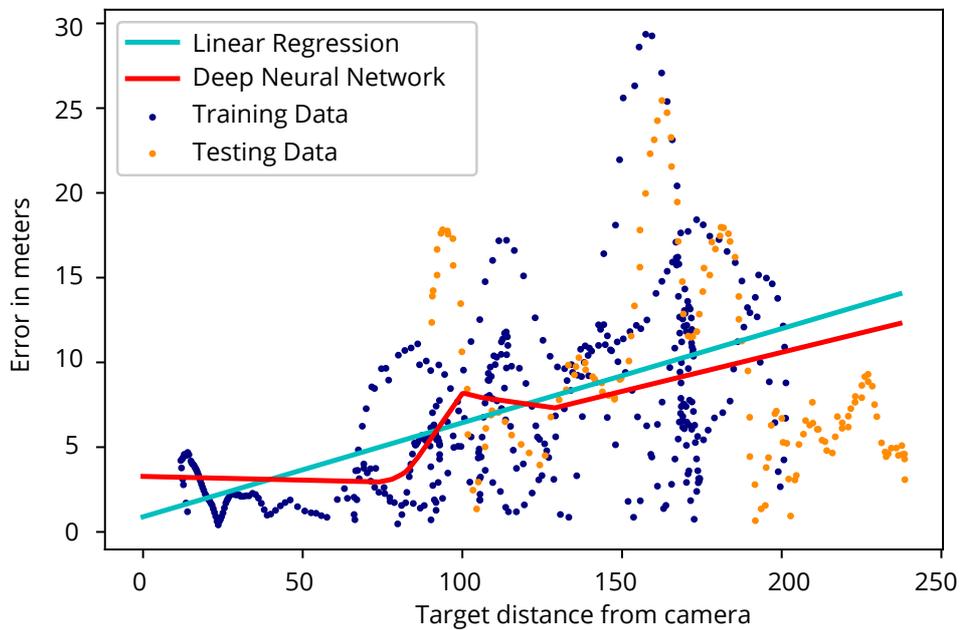}
    \caption{Visualization of the error prediction performance on the error prediction validation set. The linear regression show the average error is 5.55m per 100m distance from camera. The distance of the target vessel from the camera, against the error in meters produced by the predictions is plotted, with the training data points in blue and validation in orange. The learned regression from the Linear Regression modelling (light-blue) and the Deep Neural Network (red) is showed.}
    \label{fig:deepLearningResults}
\end{figure*}

As shown in Figures~\ref{fig:errorDistribution} and~\ref{fig:deepLearningResults}, the error performed by the model is below 20 meters for 80\% of the validation data points. The only outlying points are the ones belonging to Video 17, reaching important errors at high distances. Some of these points represent the model detecting another vessel, which was misleading these predictions. However, it can be noted that the model can predict positions around 500 and 1000 meters with a small amount of error. Predictions in the testing dataset have the highest density well below 20 meters, even reaching a precision below a single meter at the highest distances [Figures~\ref{fig:errorDistribution}, ~\ref{fig:deepLearningResults}]. It is now possible to use all the extrapolated information to perform error prediction at the end of the pipeline. The first stage of the work predicts the position of the vessel performing on the validation set. After that, the Haversine distance between both the predicted distance and ground truth, as well as the ground truth and camera position, are computed. These two parameters are used to feed the error prediction models and are respectively called \textit{prediction error} and \textit{distance from camera}.

\begin{table}[ht]
    \centering
    \begin{tabular}{c|c}
    Method & RMSE \\
    \hline
        DNN & 4.95 \\
        Linear Regression & 6.88 \\
        SVR(Linear) & 6.96 \\
        SVR(Polynomial) & 8.35 \\
        SVR(RBF) & 6.04
    \end{tabular}
    \caption{The Root Mean Squared Error - RMSE - in meters from the actual predicted point}
    \label{tab:predictionResults}
\end{table}

The first concern is, once again, the dataset. A plausible solution is to split the previous validation set according to Videos, excluding video 17. By including video 17, the predicted distance from the model would only be uncontrolled noise to our dataset, without adding any value to the scope. The training set comprises 429 data points belonging to Videos 14, 15, and 16, while the testing set has 143 points from Video 13. SVR adopts an approach similar to Support Vector Machines as Large Margin Classifiers ~\cite{cortes1995support}, with the usage of \textit{kernel tricks}~\cite{aizerman67theoretical} to create Nonlinear classifiers. The quality of the hyperplane is determined by the points falling inside the decision boundary. The results presented by Table ~\ref{tab:predictionResults} show that a simple network, with only two hidden layers and trained for 5 epochs, can reach a Root Mean Squared Error between 4 and 5 meters, using a batch size of 32 data points. The Neural Network performance is also shown by Figure~\ref{fig:deepLearningResults}. The SVR, using the Radial Basis Function kernel to approximate the nonlinear behavior, was initially prone to overfitting, achieving worse results than the Linear kernel on the testing set. To overcome this undesirable situation, the allowed margin error was increased in the training set. This allowed us to generalize better than the Linear Regression and SVR models.
The final linear regression learned is:
\begin{equation}
    \tilde{e} = 0.0555 d\,.
\end{equation}
This means that the accuracy of the pipeline is $\SI{5.55}{\meter}$ per $\SI{100}{\meter}$ distance. More data points and further experimentation would, of course, strengthen the inference capabilities of these models. Error prediction can play a significant role in the pipeline depending on the use case where this methodology could be applied. In fact, external decision-making systems could opt for a certain action or verification, depending on the detected distance of the object. 

\section{\uppercase{Discussion and Future Work}}

The results in table \ref{tab:accuracyComparison} show that our model reaches state-of-the-art precision when compared with other methodologies. However, comparisons are imprecise because of many factors like the angle of view, resolution, camera position, or target object that could influence the result. Our approach efficiently exploits GPS data with a fast and reliable inference for closer distances.

We can figure different applications in both small and large scenarios. For example, tracking a car in a parking lot where the camera is close and the accuracy should be high. A larger use-case might be keeping track of traffic in a container ship port or airport, where the camera is further away, but the objects are larger, and in this way, this approach maintains the same relative accuracy. This method could also be easily extended for cases with $z_w \neq 0$, as long as there is a clear mapping from $(x_w, y_w) \rightarrow z_w$ and $(x_p, y_p) \rightarrow z_p$.

As stated, the average error is 5.55m per 100m distance, and the network can track the ship within a sub-pixel resolution, which is crucial for distance reliability. The accuracy could be further improved by using a steeper camera angle and more training data. The performance in terms of FPS could be improved by adopting TensorRT network definition APIs and 8-bit inference~\cite{8bit}. A particular challenge for the network is to learn to distinguish the vessel from other ships like jet-skis or sailboats. This leads to the problem that multiple ships may be detected. A style detector, as presented in~\cite{wojke2017simple}, could be used to differentiate between different ships. When the object moves quickly, or the data point resolution is poor, the algorithm struggles with inferring the trajectories. For ships, one data point every second is recommended. 

In edge cases, for example, when the ship leaves the frame, the whole pipeline is susceptible to the hyper-parameters chosen, for example, how long a trajectory lasts without new positions. When the ship is at a reasonable distance and well visible in more standard situations, the algorithm is stable and precise. Since our approach requires a single frame every second, it can easily be run on resource-constrained devices at the edge achieving a low inference time while being able to track a specific vessel as well as multiple different ones, depending on the needs and training data available.

%----------------------------------------------------------------------------------------
%	REFERENCE LIST
%----------------------------------------------------------------------------------------
\bibliographystyle{apalike}
{\small
\bibliography{sources}}

\begin{thebibliography}{}

\bibitem[Aizerman et~al., 1964]{aizerman67theoretical}
Aizerman, M.~A., Braverman, E.~A., and Rozonoer, L. (1964).
\newblock Theoretical foundations of the potential function method in pattern
  recognition learning.
\newblock {\em Automation and Remote Control}, 25:821--837.

\bibitem[Ali and Hussein, 2016]{7759904}
Ali, A.~A. and Hussein, H.~A. (2016).
\newblock Distance estimation and vehicle position detection based on monocular
  camera.
\newblock In {\em 2016 Al-Sadeq International Conference on Multidisciplinary
  in IT and Communication Science and Applications (AIC-MITCSA)}, pages 1--4.

\bibitem[Bhoi, 2019]{Amlaan2019}
Bhoi, A. (2019).
\newblock Monocular depth estimation: A survey.

\bibitem[Cortes and Vapnik, 1995]{cortes1995support}
Cortes, C. and Vapnik, V. (1995).
\newblock Support vector networks.
\newblock {\em Machine Learning}, 20:273--297.

\bibitem[Dhall, 2018]{DBLP:journals/corr/abs-1809-10548}
Dhall, A. (2018).
\newblock Real-time 3d pose estimation with a monocular camera using deep
  learning and object priors on an autonomous racecar.
\newblock {\em CoRR}, abs/1809.10548.

\bibitem[Duda and Hart, 1972]{10.1145/361237.361242}
Duda, R.~O. and Hart, P.~E. (1972).
\newblock Use of the hough transformation to detect lines and curves in
  pictures.
\newblock {\em Commun. ACM}, 15(1):11–15.

\bibitem[Hu et~al., 2019]{hu2019joint}
Hu, H.-N., Cai, Q.-Z., Wang, D., Lin, J., Sun, M., Krähenbühl, P., Darrell,
  T., and Yu, F. (2019).
\newblock Joint monocular 3d vehicle detection and tracking.

\bibitem[Jiao et~al., 2019]{8825470}
Jiao, L., Zhang, F., Liu, F., Yang, S., Li, L., Feng, Z., and Qu, R. (2019).
\newblock A survey of deep learning-based object detection.
\newblock {\em IEEE Access}, 7:128837--128868.

\bibitem[Jocher et~al., 2020]{glenn_jocher_2020_4154370}
Jocher, G., Stoken, A., Borovec, J., NanoCode012, ChristopherSTAN, Changyu, L.,
  Laughing, tkianai, Hogan, A., lorenzomammana, yxNONG, AlexWang1900, Diaconu,
  L., Marc, wanghaoyang0106, ml5ah, Doug, Ingham, F., Frederik, Guilhen,
  Hatovix, Poznanski, J., Fang, J., Yu, L., changyu98, Wang, M., Gupta, N.,
  Akhtar, O., PetrDvoracek, and Rai, P. (2020).
\newblock {ultralytics/yolov5: v3.1 - Bug Fixes and Performance Improvements}.

\bibitem[Kalman, 1960]{Kalman1960}
Kalman, R.~E. (1960).
\newblock A new approach to linear filtering and prediction problems.
\newblock {\em Transactions of the ASME--Journal of Basic Engineering},
  82(Series D):35--45.

\bibitem[Leira et~al., 2017]{leira2017object}
Leira, F.~S., Helgesen, H.~H., Johansen, T.~A., and Fossen, T.~I. (2017).
\newblock Object detection, recognition, and tracking from uavs using a thermal
  camera.
\newblock {\em Journal of Field Robotics}.

\bibitem[Lin et~al., 2015]{lin2015microsoft}
Lin, T.-Y., Maire, M., Belongie, S., Bourdev, L., Girshick, R., Hays, J.,
  Perona, P., Ramanan, D., Zitnick, C.~L., and Dollár, P. (2015).
\newblock Microsoft coco: Common objects in context.

\bibitem[{Liu} et~al., 2010]{Liu2010}
{Liu}, H., {Wang}, C., {Lu}, H., and {Yang}, W. (2010).
\newblock Outdoor camera calibration method for a gps camera based surveillance
  system.
\newblock In {\em 2010 IEEE International Conference on Industrial Technology},
  pages 263--267.

\bibitem[M, 2010]{MMGM2018}
M, M. M.~G. (2010).
\newblock Landmark based shortest path detection by using a* and haversine
  formula.
\newblock {\em International Journal on Recent and Innovation Trends in
  Computing and Communication}, 6(7):98--101.

\bibitem[Migacz, 2017]{8bit}
Migacz, S. (2017).
\newblock 8-bit inference with tensorrt.
\newblock GPU Technology Conference.

\bibitem[Moritz, 2000]{Moritz2000}
Moritz, H. (2000).
\newblock Geodetic reference system 1980.
\newblock {\em Journal of Geodesy}, 74(1):128--133.

\bibitem[Sanyal et~al., 2020]{9071439}
Sanyal, S., Bhushan, S., and Sivayazi, K. (2020).
\newblock Detection and location estimation of object in unmanned aerial
  vehicle using single camera and gps.
\newblock In {\em 2020 First International Conference on Power, Control and
  Computing Technologies (ICPC2T)}, pages 73--78.

\bibitem[Snyder, 1987]{Snyder1987}
Snyder, J.~P. (1987).
\newblock Map projections: A working manual.
\newblock Technical report, Washington, D.C.

\bibitem[Tall, 2020]{Tall2020}
Tall, M.~H. (2020).
\newblock Ai tracks at sea.
\newblock \url{https://www.challenge.gov/challenge/AI-tracks-at-sea/}.

\bibitem[Tan et~al., 2020]{tan2020efficientdet}
Tan, M., Pang, R., and Le, Q.~V. (2020).
\newblock Efficientdet: Scalable and efficient object detection.

\bibitem[Wojke et~al., 2017]{wojke2017simple}
Wojke, N., Bewley, A., and Paulus, D. (2017).
\newblock Simple online and realtime tracking with a deep association metric.

\end{thebibliography}

%----------------------------------------------------------------------------------------

\end{document}